\documentclass[runningheads]{llncs}
\usepackage{graphicx}
\usepackage{multirow}
\usepackage{amssymb}
\usepackage{subfig}
\usepackage[table]{xcolor}
\begin{document}
\title{ \emph{ECOL}: \underline{E}arly Det\underline{ec}tion of \underline{CO}VID \underline{L}ies Using Content, Prior Knowledge and Source Information}
%
%
\author{Ipek Baris \and
Zeyd Boukhers}
\authorrunning{Baris and Boukhers}
%
\institute{Institute WeST, University of Koblenz-Landau, Germany \\
\email{\{ibaris,boukhers\}@uni-koblenz.de}}
\maketitle              
\begin{abstract}
Social media platforms are vulnerable to fake news dissemination, which causes negative consequences such as panic and wrong medication in the healthcare domain. Therefore, it is important to automatically detect fake news in an early stage before they get widely spread. This paper analyzes the impact of incorporating content information, prior knowledge, and credibility of sources into models for the early detection of fake news. We propose a framework modeling those features by using BERT language model and external sources, namely Simple English Wikipedia and source reliability tags. The conducted experiments on CONSTRAINT datasets demonstrated the benefit of integrating these features for the early detection of fake news in the healthcare domain. 

\keywords{Fake news detection \and Deep learning \and Prior knowledge}
\end{abstract}
\section{Introduction}
Social media is replacing traditional media as a source of information due to the ease of access, fast sharing, and the freedom to create content. However, social media is also responsible for spreading the massive amount of fake news \cite{VosoughiSpread2018}. Fake news propagation can manipulate significant events such as political elections or severely damage the society during crisis \cite{LazerScience2018}. For example, a rumor that initially occurred in a UK tabloid paper claimed that neat alcohol could cure COVID-19. As a consequence of the spread of this rumor, hundreds of Iranians have lost their lives due to alcohol poisoning\footnote{\url{https://www.independent.co.uk/news/world/middle-east/iran-coronavirus-methanol-drink-cure-deaths-fake-a9429956.html}}. Therefore, it is crucial to detect potentially false claims early before they reach large audiences and cause damage.

\begin{figure*}[!ht]
    \centering
    \includegraphics[width=\textwidth]{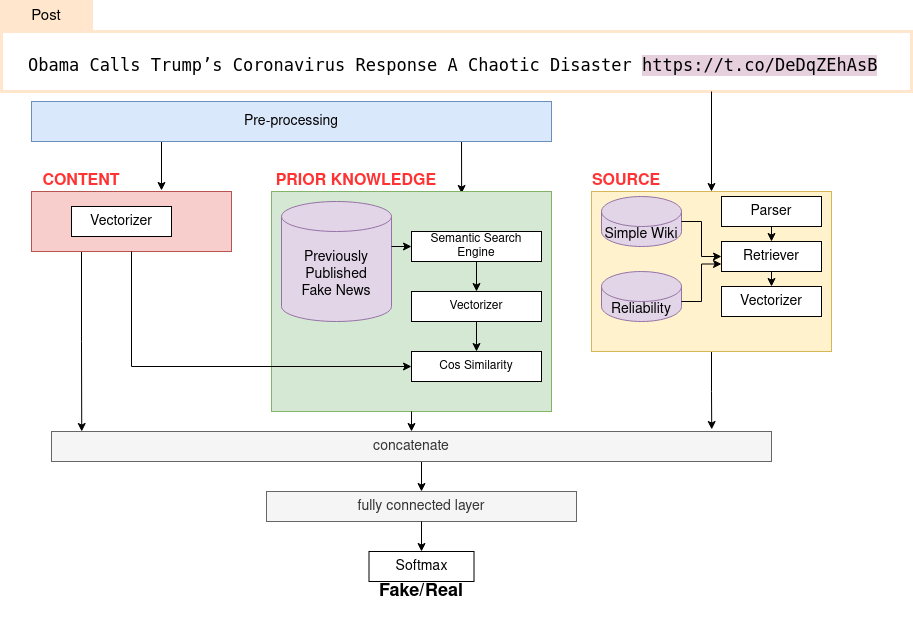}
    \caption{System architecture of \emph{ECOL} when metadata is unavailable.}
    \label{fig:arch}
\end{figure*}

Since the U.S presidential elections in 2016, tremendous efforts have been devoted by the research community to automate fake news detection. Most prior studies rely on leveraging propagation information, user engagement and content of news/social media posts \cite{survey-zhou,ShuSWTL17,0001DYL020}. However, the methods relying on propagation information \cite{YuanMZHH20} and/or on user engagement \cite{LiZS19,CastilloMP11,BarisSS19,abs-2004-01732} are not applicable for detecting fake news at an early stage since they are only available when the news starts disseminating. The methods solely based on content (e.g \cite{BarisSS19,Wang17}) could be misguided by claims that require additional context for their interpretation. For instance, the post in Figure \ref{fig:arch} may sound very plausible for readers who know the relationship between the politicians mentioned in the post. However, the source is a satiric website which indicates that the post is fake.

Commonly, a lot of fake claims/news that are partially sharing similar content occur in different sources in a relatively long time-span. For example, in the healthcare domain, the most common fake claims are about unproven and alternative cures (e.g using alcohol) against diseases \cite{the2018oncology,waszak2018spread}. These types of fake news have also been observed during COVID-19 \cite{naeem2020exploration}.  The assumption is that early published claims are fact-checked and can be employed to detect later ones. Therefore, investigating previously published claims/news can provide important information in determining the truthfulness of posts \cite{HassanZACJGHJKN17,ShaarBMN20}. Specifically, encoding previously published fake news in healthcare domain as prior knowledge, content and source information could help detecting posts that would potentially be missed by solely content-based methods. 

In this paper, we investigate and analyse those intrinsic and extrinsic features to detect fake news in the healthcare domain. To this end, we propose a Neural Network model that integrates (1) a contextual representation of the news content, (2) a representation of the relationship to similar validated fake news and (3) a source representation the embeds the reliability of the source. The main contributions of this paper are as follows:
\begin{itemize}
    \item We introduced a classification framework that models prior knowledge, content and source information by using BERT\cite{DevlinCLT19} and reliability tags to predict the truthfulness of social media posts in the healthcare domain. We share our source code and trained models on Github\footnote{\url{https://github.com/isspek/FakeNewsDetectionFramework}}.
    \item To evaluate the effectiveness of proposed model, we conducted an extensive experiment on the CONSTRAINT dataset. According to the obtained F1 score, \emph{ECOL} is ranked 14 among 167 submissions at CONSTRAINT competition \cite{patwa2021overview}. 
\end{itemize}

The rest of the paper is organized as follows. Section~\ref{related_work} briefly presents related work. Section~\ref{methodology} describes the proposed framework. Section~\ref{experiments} presents the dataset, baselines, ablation models and implementation details. Section~\ref{results} presents and discusses the results of the experiment. Finally, Section~\ref{conclusion} concludes the paper and gives insights on future work.

\section{Related Work \label{related_work}}
This section presents related work of textual content-based methods and approaches using external information to assess truthfulness.
\subsection{Textual Content based Fake News Detection on Social Media}
Textual content-based methods for fake news detection on social media~\cite{survey-zhou,ShuSWTL17,0001DYL020} vary from traditional machine learning models~\cite{CastilloMP11,KwonCJCW13,zhou2020fake} to neural networks~\cite{PetersNIGCLZ18,BarisSS19,DevlinCLT19}. For instance, the methods~\cite{CastilloMP11,KwonCJCW13} leverage features such as the sentiment and metadata information (e.g. replies, likes of social media posts) in addition to statistics derived from both post and metadata. Zhou et al. \cite{zhou2020fake} investigate the features derived by the theories in social and forensic psychology. As examples of neural network models, CNN \cite{Wang17}, RNN \cite{ChenLYZ18} and most recently context-aware language models such as ELMo \cite{PetersNIGCLZ18,BarisSS19} and BERT \cite{FajcikSB19,DevlinCLT19} have also been used. While CNN, RNN models ignore the context information, BERT and ELMo can learn the different meanings of the words depending on the context. Among context-aware language models, BERT has shown state-of-art results in many NLP tasks \cite{DevlinCLT19}. Therefore, in our study, we encode content information with BERT. 

\subsection{Extrinsic Features for Determining Truthfulness of Claims}
As an extrinsic feature, encoding the top $N$ relevant evidence pages retrieved by commercial search engines (e.g. Google) has been a widely preferred approach~\cite{abs-1811-10971,AugensteinLWLHH19,LiZ20} to determine the truthfulness of claims or posts. However, Augenstein et al.~\cite{AugensteinLWLHH19} stated that this method has a drawback of affecting veracity assessments when the results change over time. Firstly, this drawback could prevent reproducible results. Secondly, it would not be applicable to evaluate the posts which cannot be supported or denied with evidences when initially occurred.

Another extrinsic feature is leveraging the information of previously analyzed claims. Claim similarity between previously fact-checked claims in the political domain has been studied as part of fact-checking system \cite{HassanZACJGHJKN17,DenauxG20,abs-2002-06585} or as an information retrieval task \cite{ShaarBMN20}. Those studies aim to find claims or posts reporting about the same event. However, we aim to learn the similarity of the posts with previously detected fake news in the healthcare domain, not necessarily reporting about the same event.

Lastly, the credibility of user-profiles \cite{LiZS19,YuanMZHH20} and source websites \cite{MensioA19,abs-1809-00494,NorregaardHA19,NELA2019} are also strong extrinsic features for determining truthfulness at an early stage \cite{abs-2008-04374}. To detect rumors and fake news on social media, Yuan et al. \cite{YuanMZHH20} used user credibility as weak signals in their graph-based neural network model. Li et al. \cite{LiZS19} combined the credibility of users with post features and post embeddings for rumor detection. These two studies require propagation and metadata information to encode the aforementioned features. Other studies~\cite{MensioA19,NorregaardHA19,NELA2019} focus on determining the credibility of sources that mostly report political news. Using only the credibility information of political news sources could be limited. Therefore, we leverage the content of the Simple English Wikipedia and source credibility information by Gruppi et al. \cite{NELA2019} in our study.

In summary, \emph{ECOL} utilizes the content of the post, its similarity to prior knowledge, and the credibility of its embedded source URLs, to detect their truthfulness early. 

\section{\emph{ECOL} Approach \label{methodology}}
The overall architecture of \emph{ECOL} framework is illustrated in Figure~\ref{fig:arch}. Firstly, as a pre-processing step, the framework (1) lowers all cased words, (2) fixes the Unicode errors, (3) translates the text to the closest ASCII representation and (4) replaces exclusive content with tags. Specifically, URLs are replaced with with \texttt{<URL>}, emails with \texttt{<EMAIL>}, numbers with \texttt{<NUMBER>}, digits with \texttt{<DIGIT>}, phone numbers with \texttt{<PHONE>} and currency symbols with \texttt{<CUR>}. Secondly, the framework encodes (1) content information from solely posts (Section \ref{Content}), (2) relation with top 10 relevant fake news in health domain (Section \ref{PriorKnowledge}) and (3) the sources, that are embedded within the post, by concatenating reliability tags and their Simple Wiki descriptions (Section \ref{Source}). Next, it concatenates the encoded features and feeds them into a fully connected layer. Finally, a softmax layer classifies the features as fake or real to express their truthfulness. 

\subsection{Content (C)\label{Content}}
In this unit, \emph{ECOL} tries to capture the writing style of fake and true posts. To well learn the content information, we encode the texts with BERT~\cite{DevlinCLT19} which is a context-aware language model based on the transformer network and has been pre-trained on massive text corpus~\cite{vaswani2017attention}. BERT learns specific NLP tasks after fine-tuning its pre-trained models. In order to obtain the post representations, we encode the input sequences of the post with the uncased base version of pre-trained BERT by using the \texttt{transformers} library \cite{wolf-etal-2020-transformers}. Uncased base BERT consists of 12 layers and 12 attention heads and outputs 768-dimensional vectors for each word in the post. The first token of the input sequences is called \texttt{[CLS]} and indicates the classification label. The final hidden state of the \texttt{[CLS]} is used as a content representation. We tune the maximum size of texts to 128 and padded short texts.  

\subsection{Prior Knowledge (PK)\label{PriorKnowledge}}
To leverage the relation of news event with previously published and proved fake news, we encoded a post's relation with a set of similar fake news disseminated before COVID-19. Given a post as a query, an ad-hoc semantic search engine retrieves the top 10 fake news from a repository indexed with fake news in the health domain. To obtain a fake news vector (\textbf{FN}), we encoded each retrieved news with the BERT and took their average. Afterwards, we computed the \emph{cosine} similarity between \textbf{FN} and the post (\textbf{P}) as follows $\mathbf{R}=cos(\mathbf{FN}, \mathbf{P})$, where $\mathbf{R}$ is an one dimensional relatedness vector.
\\
\\
\textbf{3.2.1 Ad Hoc Search and Indexing} \\
\\
To obtain the similar fake news, we used Elasticsearch \footnote{\url{docker.elastic.co/elasticsearch/elasticsearch:7.6.1}} to retrieve  validated fake news from FakeHealth~\cite{DaiSW20} which is a dataset containing real and fake news stories and releases published in 2009 and 2018 from the health fatchecking organization \texttt{HealthNewsReview}\footnote{\url{https://www.healthnewsreview.org/}}. The dataset covers diseases such as cancer, alzheimer, etc. \\
We indexed the title and article of the news in text format and to add the ability of semantic retrieval to the search engine, we encode title and the article of news also with the pre-trained sentence-BERT~\cite{ReimersG19}. The search engine retrieves the top 10 documents whose fields match and have high cosine similarity with the query. If the number of retrieved documents is smaller than 10, the list of the documents is appended with an empty strings.

\subsection{Source (S)\label{Source}}
To encode information of the sources, for each source, we first unshortened any shortened links, such as the URL in Figure~\ref{fig:arch}. Then, we extract the name of the source (e.g \texttt{thespoof}). We retrieve then the reliability and Simple Wiki description of the source and vectorized source information by concatenating the retrieved information.   
\\
\\
\textbf{3.3.1 Reliability}\\
\\
NELA 2019 is a dataset containing 260 news sources proposed by Gruppi et al~\cite{NELA2019}. The dataset contains source reliability labels from various assessment websites such as Media Bias/Fact Check (MBFC)\footnote{\url{https://mediabiasfactcheck.com/}}, Politifact\footnote{\url{https://www.politifact.com/}}, etc. Moreover, The authors aggregated the labels from MBFC by assigning a label \emph{unreliable} to sources with low factual reporting or listed as conspiracy/pseudoscience source. Similarly, they assigned \emph{reliable} to sources with high factual reporting. We determined the source reliability by combining the aggregated labels and satire sources from MBFC. In the end, the source types that we used for reliability are \emph{reliable}, \emph{unreliable}, and \emph{satire}. We assign the \texttt{na} (not available) tag to the sources that do not occur in NELA 2019. Afterwards, we vectorize the reliability of each source with one hot encoder. For the posts that do not have URLs or the number of URLs less than 5 URLs, we append the source lists with zero vectors of a size equal to the length of the reliability tags.
\\
\\
\textbf{Simple Wiki Source Descriptions}\\
\\
The Simple English Wikipedia aims to provide access to an English encyclopedia for non-native English speakers and children. The entity descriptions of Simple English Wikipedia can be a clue for the trustworthiness of the sources. We downloaded the Simple English Wikipedia, February 2019 dump\footnote{\url{https://archive.org/details/simplewiki-20190201}} to ensure that the contents were written before COVID-19. Using the tool Wiki Extractor\footnote{\url{https://github.com/attardi/wikiextractor}}, we constructed a dictionary mapping entities to their descriptions. For simplicity, we ignored ambiguous entities that have more than one wiki pages and mapped the description of each source in the posts if a key of the dictionary exactly matched the source name. Afterwards, we encoded each source descriptions as BERT representations.

\section{Experiments \label{experiments}}

\subsection{Dataset}
The CONSTRAINT dataset~\cite{abs-2011-03327} contains fake and real news about COVID-19 in the form of social media posts. Fake news samples were collected from various fact-checking websites and tools such as Politifact, IFCN chatbot. Real news samples were collected from verified Twitter accounts and manually checked by the organizers. The dataset is split into train, dev, and test splits. Table~\ref{tab:data_stats} presents statistical details of the datasets.

\begin{table}[!ht]
    \caption{Statistical details of the task's dataset~\cite{abs-2011-03327}. \textbf{w:} post with links, \textbf{w/o:} posts without links}
    \centering
    \begin{tabular}{|l||ll||ll||l|l|l}
    \hline
         & \multicolumn{2}{l}{\textbf{Real}} & \multicolumn{1}{l}{\textbf{Fake}} && \multirow{2}{*}{\textbf{Total}} \\
         \cline{2-5}
         & \textbf{w} &  \textbf{w/o} & \textbf{w} & \textbf{w/o} &\\
         \hline
         \textbf{Train} & 2321 & 1039 & 1002  & 2058 & 6420\\ 
         \textbf{Dev} & 780 & 340 & 327 &693  & 2140\\ 
         \textbf{Test} & 779 & 341 &319 & 701 & 2140\\ 
         \hline
    \end{tabular}
    \label{tab:data_stats}
\end{table}

\subsection{Baselines}
We compared \emph{ECOL} framework against the baseline classifiers provided by the organizers, which are Support Vector Machine (SVM), Logistic Regression (LR), Gradient Boost, and Decision Trees (DT). The baseline classifiers are trained on term frequency-inverse document frequency (tf-idf) features. 

\subsection{Models}
We compare and analyse the following variations of \emph{ECOL} framework models:
\begin{itemize}
\item \textbf{C} uses solely content information as feature.
    \item \textbf{PK} uses solely prior knowledge as feature.
    \item \textbf{C\_PK} uses concatenation of content and prior knowledge as feature.
    \item \textbf{C\_S} uses concatenation of content and source as feature.
    \item \textbf{C\_PK\_S} uses concatenation of content, source and prior knowledge.
\end{itemize}

\subsection{Implementation}
We implemented \emph{ECOL} models using PyTorch Lightning\footnote{\url{https://pytorch-lightning.readthedocs.io/en/0.7.1/introduction\_guide.html}}. We trained the models with 42, 0, 36 random seeds, three epochs, and one batch size on a NVIDIA TITAN RTX 16 GB GPU.  

\section{Results and Discussion\label{results}}
We present the experimental results on development and test sets in Table~\ref{tab:results_val}. We report Precision (P), Recall (R), F1 scores per class, and accuracy, weighted P, R, and F1 scores as the models' overall performance. \textbf{C$\mu$}, \textbf{PK$\mu$}, \textbf{C\_S$\mu$}, \textbf{C\_PK\_S$\mu$} average the predictions by the models trained with 42, 36, 0 as random seeds. The other models use a random seed of 42, which gave the highest F1 scores in our experiments. We entered the CONSTRAINT shared task with three entries: an average over the three \textbf{C\_PK\_S} models and the two \textbf{C\_PK\_S} models with the highest F1 scores (random seeds are 42, 36). The best performing model with random seed 42 ranked 14 among 167 submissions \cite{patwa2021overview}.
\begin{table*}[ht]
    \caption{Precision, recall and F1-score of the baseline and proposed models, trained on the CONSTRAINT datasets. Highlighted scores indicate the highest values for each metric.}
    \centering
    \begin{tabular}{|l|l||lll||lll||llll|} 
    \hline
    \multirow{2}{*}{\textbf{Set}} & \multirow{2}{*}{\textbf{Model}} & \multicolumn{2}{l}{\textbf{Fake}} & &\multicolumn{2}{l}{\textbf{Real}} & & \multicolumn{3}{l}{\textbf{Overall}} & \\
    & &\textbf{P} & \textbf{R} & \textbf{F1} & \textbf{P} & \textbf{R} & \textbf{F1} & \textbf{Acc} & \textbf{P} & \textbf{R} & \textbf{F1}\\
    \hline
    \multirow{14}{*}{\textbf{Dev}} & \textbf{SVM} & 92.07 & 94.41 &93.22 &94.79 &92.59 &93.68 & 93.46 & 93.48 &93.46 &93.46 \\
    &\textbf{LR} & 91.07 & 94.02 &92.52 &94.39 &91.61 &92.98 & 92.76& 92.79 & 92.76 & 92.75\\
    &\textbf{GB} & 83.41 &90.20 & 86.67 &90.36 &83.66 &86.88 &86.78& 87.03 & 86.78 &86.77\\
    &\textbf{DT} & 85.53 &83.43 &84.47 &85.24 &87.14 &86.18 &85.37 &85.42 &85.37 &85.38\\\cline{2-12}
    &\textbf{C} & 98.12 & 97.06  & 97.59 & 97.35 & 98.30 & 97.82 & 97.71 & 97.72 & 97.71 & 97.71 \\
    &\textbf{PK} & 97.69  & 95.20 & 96.43 & 95.72 &  97.95 & 96.82 & 96.63 & 96.67 & 96.64 & 96.64 \\
    &\textbf{C\_PK} & \cellcolor{blue!25}\textbf{98.99}  &95.78& 97.36 &96.27 & \cellcolor{blue!25}\textbf{99.11} & 97.67 & 97.52 &97.57 &97.52 &97.53\\
    &\textbf{CS} & 98.39 & 95.98 & 97.17 & 96.42 &98.57 &97.48 & 97.34 & 97.37 & 97.34 & 97.34\\
    &\textbf{C\_PK\_S} &98.51 & \cellcolor{blue!25}\textbf{97.25} & \cellcolor{blue!25}\textbf{97.88} & \cellcolor{blue!25}\textbf{97.53} &98.66 & \cellcolor{blue!25}\textbf{98.09} & \cellcolor{blue!25}\textbf{97.99} & \cellcolor{blue!25}\textbf{98.00} & \cellcolor{blue!25}\textbf{97.99} & \cellcolor{blue!25}\textbf{97.99}\\\cline{2-12}
    &\textbf{C$\mu$} & 99.28  &  95.20  & 97.20 &  95.78 & 99.38  & 97.55 & 97.38 & 97.47 & 97.38 & 97.39 \\
    &\textbf{PK$\mu$}  & 99.17 & 93.53  & 96.27 & 94.40 &  99.29 & 96.78 & 96.54 & 96.70  & 96.54 & 96.55\\
    &\textbf{C\_PK$\mu$} & \cellcolor{blue!25}\textbf{99.49}  & 94.71 & 97.04  &  95.38 & 99.55 & 97.42 & 97.24 &  97.35 & 97.24 & 97.24 \\
    &\textbf{C\_S$\mu$}  & \cellcolor{blue!25}\textbf{99.49}  & 95.10   & 97.24 & 95.71 & 99.55 & 97.59 & 97.43 &  97.52 & 97.43 & 97.43\\
    &\textbf{C\_PK\_S$\mu$}  & 99.09  &  \cellcolor{blue!25}\textbf{96.08} & \cellcolor{blue!25}\textbf{97.56} & \cellcolor{blue!25}\textbf{96.52} & \cellcolor{blue!25}\textbf{99.20} &\cellcolor{blue!25}\textbf{97.84} & \cellcolor{blue!25}\textbf{97.71} & \cellcolor{blue!25}\textbf{97.75} & \cellcolor{blue!25}\textbf{97.71} & \cellcolor{blue!25}\textbf{97.71} \\
    \hline
    \multirow{14}{*}{\textbf{Test}} & \textbf{SVM}   & 92.20 &  93.92 & 93.05 & 94.37 & 92.77 & 93.56 & 93.32 & 93.33  & 93.32 & 93.32\\
    & \textbf{LR}   & 90.08  & 93.43  & 91.72 & 93.81 & 90.62 & 92.19 & 91.96 & 92.01 & 91.96 & 91.96 \\
    & \textbf{GB}   & 83.39 &  90.59 & 86.84 & 90.70 & 83.57 & 86.99 & 86.92 & 87.20 & 86.92 & 86.91\\
    & \textbf{DT}  & 85.39 &  84.22 &  84.80 & 85.80 & 86.88 & 86.34 &85.61  & 85.62 & 85.61 & 85.61\\
    \cline{2-12}
    & \textbf{C} & 98.21 & \cellcolor{blue!25}\textbf{96.96}  & 97.58 & 97.26 & 98.39 & 97.83 & 97.71 & 97.72 & 97.71  & 97.71\\
    & \textbf{PK}   & 97.78 & 95.20  & 96.47 & 95.73 & 98.04 & 96.87 & 96.68 & 96.72 & 96.68 & 96.68 \\
    & \textbf{C\_PK}   & \cellcolor{blue!25}\textbf{99.29} & 95.59  & 97.40 & 96.11 & \cellcolor{blue!25}\textbf{99.38} & 97.72 & 97.57 & 97.64 & 97.57 & 97.57\\
    & \textbf{CS}   & 98.69 & 96.08 &  97.37 & 96.51 & 98.84 & 97.66 & 97.52 & 97.56 & 97.52 & 97.53 \\
    & \textbf{C\_PK\_S}   & 99.10 & \cellcolor{blue!25}\textbf{96.96}  & \cellcolor{blue!25}\textbf{98.02} & \cellcolor{blue!25}\textbf{97.29} & 99.20 & \cellcolor{blue!25}\textbf{98.23} & \cellcolor{blue!25}\textbf{98.13} & \cellcolor{blue!25}\textbf{98.15} & \cellcolor{blue!25}\textbf{98.13} & \cellcolor{blue!25}\textbf{98.13} \\
    \cline{2-12}
    & \textbf{C$\mu$}  & 99.49 & 94.80  & 97.09 & 95.46 & 99.55 & 97.47 & 97.29 & 97.40 & 97.30 & 97.29\\
    & \textbf{PK$\mu$}& 98.56 & 94.02  & 96.24 & 94.77 & 98.75 & 96.72& 96.50 & 96.60 & 96.50 & 96.50 \\
    & \textbf{C\_PK$\mu$}  & 99.38 & 93.82   & 96.52 &94.65 & 99.46 & 97.00 & 96.78  & 96.93 & 96.78 & 96.78\\
    & \textbf{C\_S$\mu$}  & \cellcolor{blue!25}\textbf{99.79} & 94.90  & 97.29 & 95.56 & \cellcolor{blue!25}\textbf{99.82}  & 97.64 & 97.48 & 97.59  & 97.48 & 97.48\\
    & \textbf{C\_PK\_S$\mu$}  & 99.59 & \cellcolor{blue!25}\textbf{95.29}  & \cellcolor{blue!25}\textbf{97.39} & \cellcolor{blue!25}\textbf{95.88} & 99.64 & \cellcolor{blue!25}\textbf{97.72} & \cellcolor{blue!25}\textbf{97.57} & \cellcolor{blue!25}\textbf{97.66} & \cellcolor{blue!25}\textbf{97.57}  & \cellcolor{blue!25}\textbf{97.57}\\
    \hline
    \end{tabular}
    \label{tab:results_val}
\end{table*}

\begin{figure}[!ht]
\centering
\subfloat[Fake news posts with and without links]{\includegraphics[width=0.9\textwidth, height=0.3\textheight]{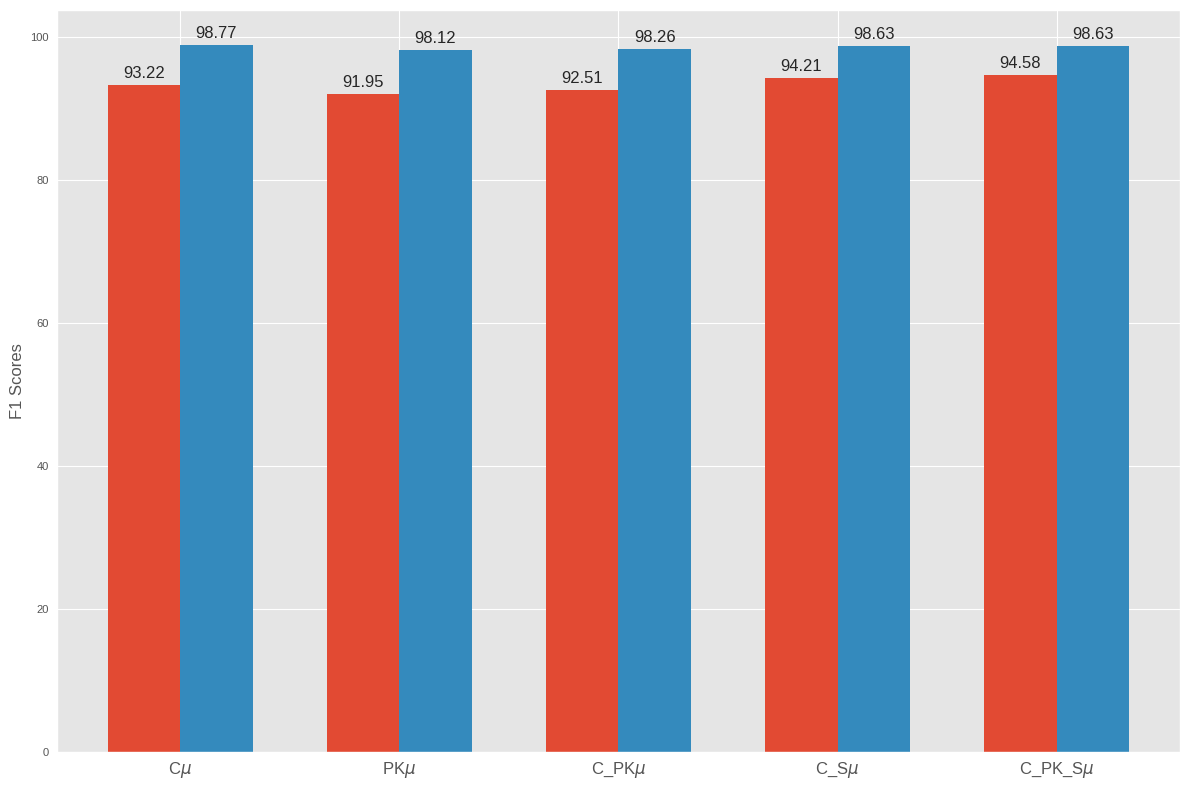}\label{Fig-a}}\hfill
\subfloat[Real news with and without links]{\includegraphics[width=0.9\textwidth, height=0.3\textheight]{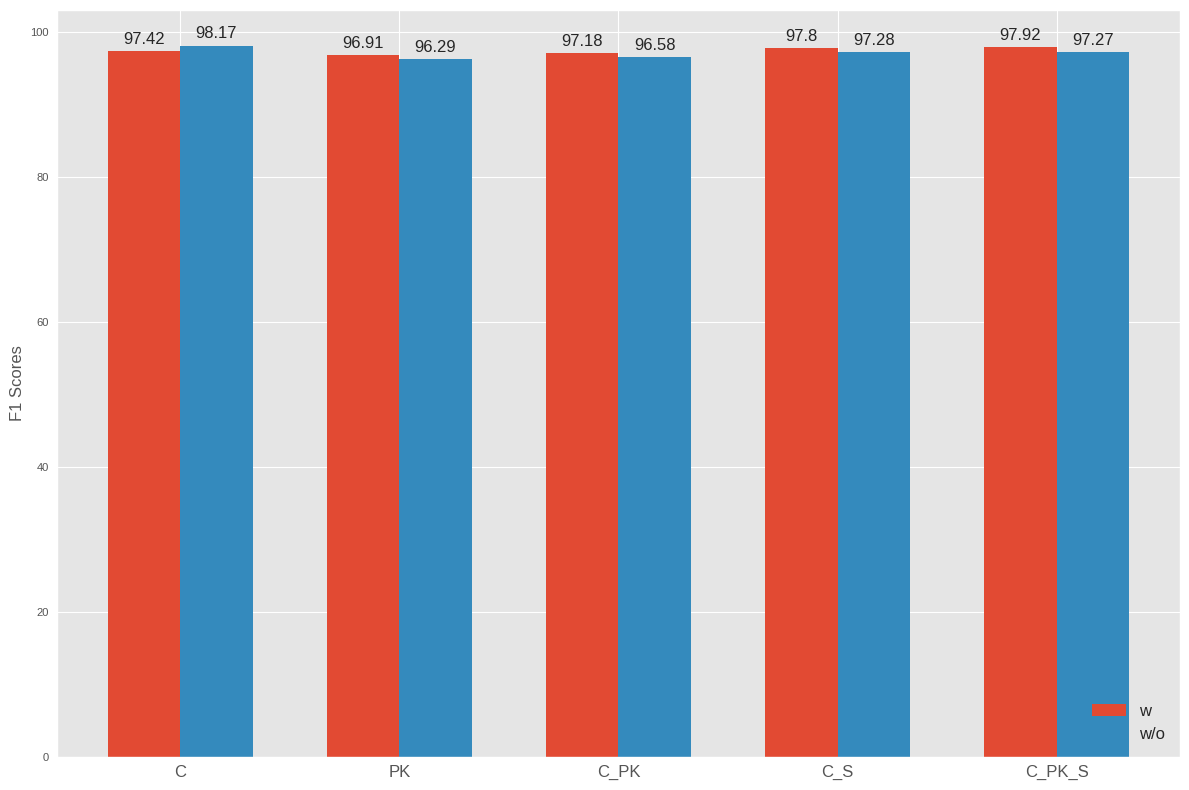}\label{Fig-b}}
\caption{F1 Scores of the $\mu$ models when predicting posts with (blue color) and without links (red color).} 
\label{fig:performance_links} 
\end{figure}

By applying a T-test at the 0.01 significance level, we observe that the proposed models significantly outperformed the baselines. When we compare the content-based models (\textbf{C} and \textbf{C$\mu$}) with the other proposed models, we first see that the prior knowledge and content information (\textbf{C\_PK}) complements the source information (\textbf{C\_S}). Moreover, the prior knowledge and content information helps to identify false news, but source information helps in identifying real news. Therefore, among the proposed models, \textbf{C\_PK\_S} and \textbf{C\_PK\_S$\mu$} achieve the highest F1 scores by balancing the predictions towards real and fake news. However, the improvement is not significant. For instance, \textbf{C} misclassified only 49 samples while the \textbf{C\_PK\_S}, classified 9 samples more correctly. 

\textbf{PK} and \textbf{C\_PK} which incorporate the prior knowledge are the least successful models among the proposed models. We observed that the indexing method for the retrieval unit yields false-positive predictions. However, the models also outperformed the official baselines which implies that prior knowledge could be useful for fake news detection. For better healthcare retrieval, we plan to improve the indexing schema with the semantic concepts that define the health claim types such as treatment, alternative medicine. 

We also analyzed how the presence of links in posts change the model predictions and present F1 scores of the models by grouping them into posts with and without links in Figure~\ref{fig:performance_links}. The presence of links in fake news drastically degrades the performance of the models (Figure~\ref{Fig-a}). For example, while \textbf{C\_S$\mu$} scores the posts with the links as 98.77, its F1 score is reduced to 93.22. However, encoding source information into the models (\textbf{C\_PK\_S$\mu$} and \textbf{C\_S$\mu$}) improves identifying fake news posts with links. When we analyse the links, we see that they are Twitter accounts, medical websites and delete links that are not present in Simple English Wikipedia nor reliability dictionary. However, we found some samples in the test set, which have links to a fact-checking website (Politifact) but were annotated as \texttt{fake}, potentially yielding false predictions.

As seen in Figure~\ref{Fig-b}, the content information is the only key feature for identifying real news. Prior knowledge and source information could not improve the prediction of real news posts with links. When we analyze real news posts that were misclassified by the models, we see that although the posts are written in reporting language, they also contain judgemental language. For example, \texttt{Coronavirus: Donald Trump ignores COVID-19 rules with 'reckless and selfish' indoor rally [URL]} might confuse the models by combining the two language types. 

\section{Conclusion\label{conclusion}}
In this paper, we presented a promising framework for the early detection of fake news on social media. The framework encodes content, prior knowledge, and credibility of sources from the URL links in the posts. We analyzed the impact of each encoded information on the models to detect fake news in the healthcare domain. We observed that using three perspectives could lead to precisely distinguish between fake and real news. In future work, we will improve the source linking by using structured data such as Wikidata \cite{vrandevcic2014wikidata} in order to encode more source knowledge. For a better retrieval in the healthcare domain, we plan also to index prior knowledge by categorizing it into semantic concepts such as cure, treatment and symptoms. 
\bibliographystyle{splncs04}
\bibliography{paper}
\end{document}